\documentclass[letterpaper]{article} 
\usepackage{aaai2027}  
\usepackage[hyphens]{url}  
\usepackage{graphicx} 
\usepackage{natbib}  
\usepackage{caption} 
\usepackage{algorithm}
\usepackage{algorithmic}
\usepackage{newfloat}
\usepackage{listings}
\DeclareCaptionStyle{ruled}{labelfont=normalfont,labelsep=colon,strut=off} 
\floatstyle{ruled}
\newfloat{listing}{tb}{lst}{}
\floatname{listing}{Listing}

\usepackage{booktabs}

\usepackage{booktabs}
\usepackage{multirow}
\usepackage[table]{xcolor}

\usepackage{makecell}
\usepackage{pifont}

\newcommand{\xmark}{\ding{55}}

\usepackage{algorithm}
\usepackage{algorithmic}

\usepackage{amsmath}
\usepackage{amssymb}

\title{Summarize Before Grounding: Query-Guided Chunk Condensation for  Long-Video Temporal Grounding}

\author{
    Nanxing Hu\textsuperscript{\rm 1},
    Xiaoyue Duan,
    Qiwei Yan\textsuperscript{\rm 2},
    Kailin Lyu\textsuperscript{\rm 3},
    Jinchao Zhang\corresponding,
    Guoliang Kang\textsuperscript{\rm 1}\corresponding
}

\affiliations{
    \textsuperscript{\rm 1}Beihang University\\
    \textsuperscript{\rm 2}University of the Chinese Academy of Sciences\\
    \textsuperscript{\rm 3}Institute of automation, Chinese academy of science
}

\begin{document}

\maketitle

\begin{abstract}

Video temporal grounding (VTG) aims to localize the video interval corresponding to a language query. 
Recent large vision-language models (LVLMs) show great potential in solving such a multi-modal reasoning task.
However, long videos often contain large amounts of redundant information that disturbs LVLMs to mine query-relevant evidence. 
Instead of dense frame sampling which incurs prohibitive training memory, previous reinforcement learning with verifiable rewards (RLVR) works typically utilize sparse sampling, which makes training feasible but may miss critical evidence. 
In this paper, we propose a ``summarize before grounding'' framework (named ``SumGround'') for long-video temporal grounding. The key of SumGround is to perform query-guided chunk condensation to aggregate and retrieve query-relevant evidence. 
Specifically, we split the video into several chunks and perform two-level chunk condensation.
First, we introduce query-guided latent summaries, which is represented as KV states of query-guided prompts, to compress redundant visual tokens into compact query-relevant chunk summaries. 
Furthermore, we design an associative summary retrieval scheme to rank and select  chunk summaries that are most likely to contain the event interval. 
Both query-guided latent summary and associative summary retrieval schemes are enabled by RLVR.
To reduce memory consumption, we propose a length-aware gradient gating module to selectively stop gradient back-propagated to visual tokens.
Extensive experiments demonstrate that SumGround performs favorably against previous state-of-the-art methods across multiple downstream datasets, with remarkable gains on long videos.
\end{abstract}


\begin{figure*}[t]
    \centering
    \includegraphics[width=\linewidth]{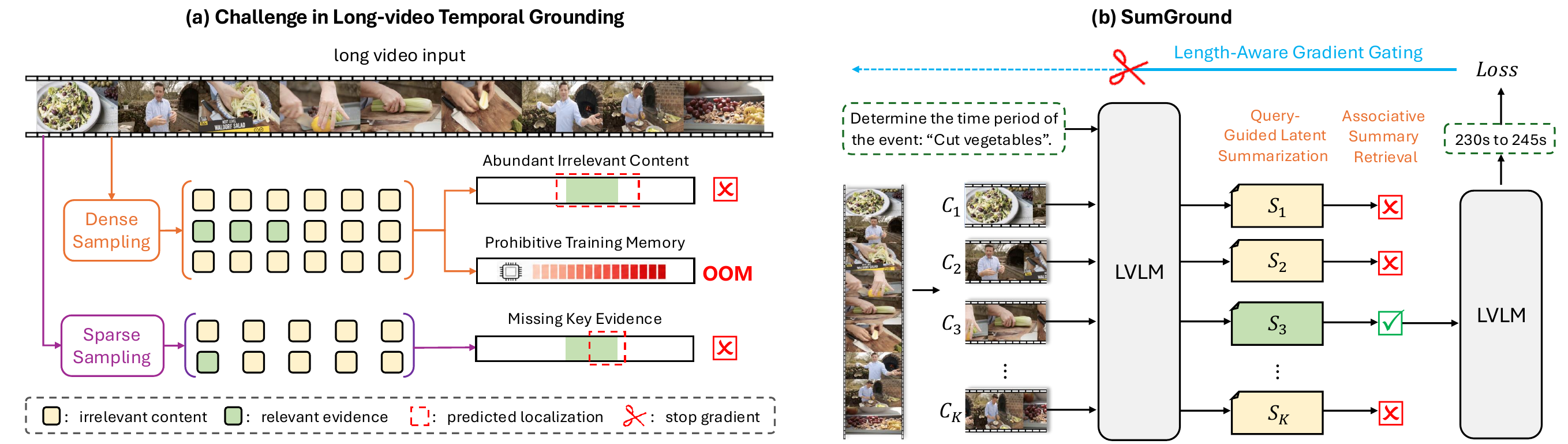}
    \caption{
    Overview of the motivation and SumGround.
    (a) Dense sampling retains fine-grained evidence but also introduces extensive query-irrelevant content and prohibitive training memory, whereas sparse sampling may miss key evidence.
    (b) SumGround processes the video chunk by chunk, compresses query-relevant evidence into sequential summary states, and retains query-associated summaries for final grounding.
    We gate visual-token gradients according to video length while continuing to optimize summary accumulation, retrieval, and grounding.
    }
    \label{fig:overview}
\end{figure*}

\section{Introduction}

Video temporal grounding (VTG) aims to localize the video interval corresponding to a language query~\cite{gao2017tall,zhang2023temporal}. 
In recent years, large vision-language models (LVLMs), due to their superior multi-modal reasoning capability, have been widely adopted to solve the VTG problem~\cite{huang2024vtimellm,wang2024hawkeye}.
However, for a long video that is quite common in practice, the queried event may occupy only a small portion, which means that a large amount of content is redundant. The redundant information may mislead LVLM, resulting in imprecise grounding boundary predictions.


Reinforcement learning with verifiable rewards (RLVR) has been widely adopted to post-train a LVLM to improve its VTG performance.
Existing RLVR-based VTG methods typically rely on sparse frame sampling because retaining dense visual tokens and their gradients over long videos is prohibitively expensive.
As illustrated in Figure~\ref{fig:overview}(a), sparse sampling reduces the number of input frames (and thus reduces the number of tokens consumed) but may miss critical evidence relevant to query.
Dense sampling covers query-relevant content, but also contains massive redundant information.
Moreover, retaining all visual embeddings and their gradients in RLVR incurs prohibitively high training memory cost.
Therefore, the key challenge is to preserve query-relevant evidence, filter out irrelevant information, and efficiently optimize the resulting long-range representation for VTG.

In this paper, we propose a ``summarize before grounding'' framework (named
``SumGround'') for long-video temporal grounding.
The key of SumGround is to perform query-guided chunk condensation to aggregate and retrieve query-relevant evidence.
As shown in Figure~\ref{fig:overview}(b), we perform two-level context condensation.
First, we split the video into several chunks and process them sequentially through multiple prefill passes. 
For each prefill, rather than generating a textual summary or introducing additional modules, we propose using query-guided latent summary, which is represented as KV states of query-guided prompts, to compress redundant visual tokens into compact query-relevant chunk summaries. 
The latent summary of the current chunk attends to current chunk visual tokens and previous chunk summaries. 
We store the KV states of summary tokens instead of the KV states of visual tokens.
Furthermore, we design an associative summary retrieval scheme to rank and select chunk summaries that are most likely to contain the event interval.
In detail, during the chunk-wise prefill stage, we utilize the predicted logits of the query-association prompt to judge whether the current chunk is associated with the query.
After all video chunks are processed, associative summary retrieval retains query-associated summaries and filters out unrelated ones, constructing a concentrated context for precise grounding.

For training, both query-guided latent summary and associative summary retrieval schemes are enabled by RLVR.  
We propose a length-aware gradient gating module to reduce memory consumption.
Specifically, our gradient gating strategy keeps the forward process unchanged and selectively stops gradients back-propagated to visual tokens based on the input video length.
For short videos, visual-token gradients are retained, allowing the model to learn precise visual token hidden states and the vision-to-summary capability. 
For long videos, visual-token gradients are blocked to save memory, while the model continues to learn how summaries are accumulated, retrieved, and exploited for temporal grounding over long temporal contexts.

We instantiate RLVR with Group Relative Policy Optimization (GRPO)~\cite{shao2024deepseekmath}, with separate rewards for associative summary retrieval and temporal grounding.
For experiments, we train LVLM on constructed training set and perform zero-shot testing on four typical VTG benchmarks, \emph{i.e.,} Charades-STA~\citep{gao2017tall}, ActivityNet-Captions~\citep{krishna2017dense}, TACoS~\citep{regneri2013grounding}, and Ego4D-NLQ~\citep{grauman2022ego4d}.
Experiments show that SumGround performs favorably against previous methods, with remarkable gains on the long videos.
Ablations verify the effectiveness of key components of our method.

In a nutshell, our contributions are summarized as follows
\begin{itemize}
    \item We propose a novel framework, \emph{i.e.,} ``summarize before grounding'', to perform query-guided chunk condensation to aggregate and retrieve query-relevant evidence for long-video temporal grounding. Our framework introduces no additional parameterized modules and can be trained end-to-end by RLVR.
    

    \item We propose two-level chunk condensation techniques, including query-guided latent summary and associative summary retrieval. For training, we design GRPO rewards respectively for associative summary retrieval and temporal grounding. Further, we propose a length-aware gradient gating module to reduce memory consumption.

    \item 
    Experiments demonstrate that our method performs favorably against previous state-of-the-art methods, especially on long videos. Ablations verify the effectiveness of the key components of our framework.
\end{itemize}

\begin{figure*}[t]
    \centering
    \includegraphics[width=\linewidth]{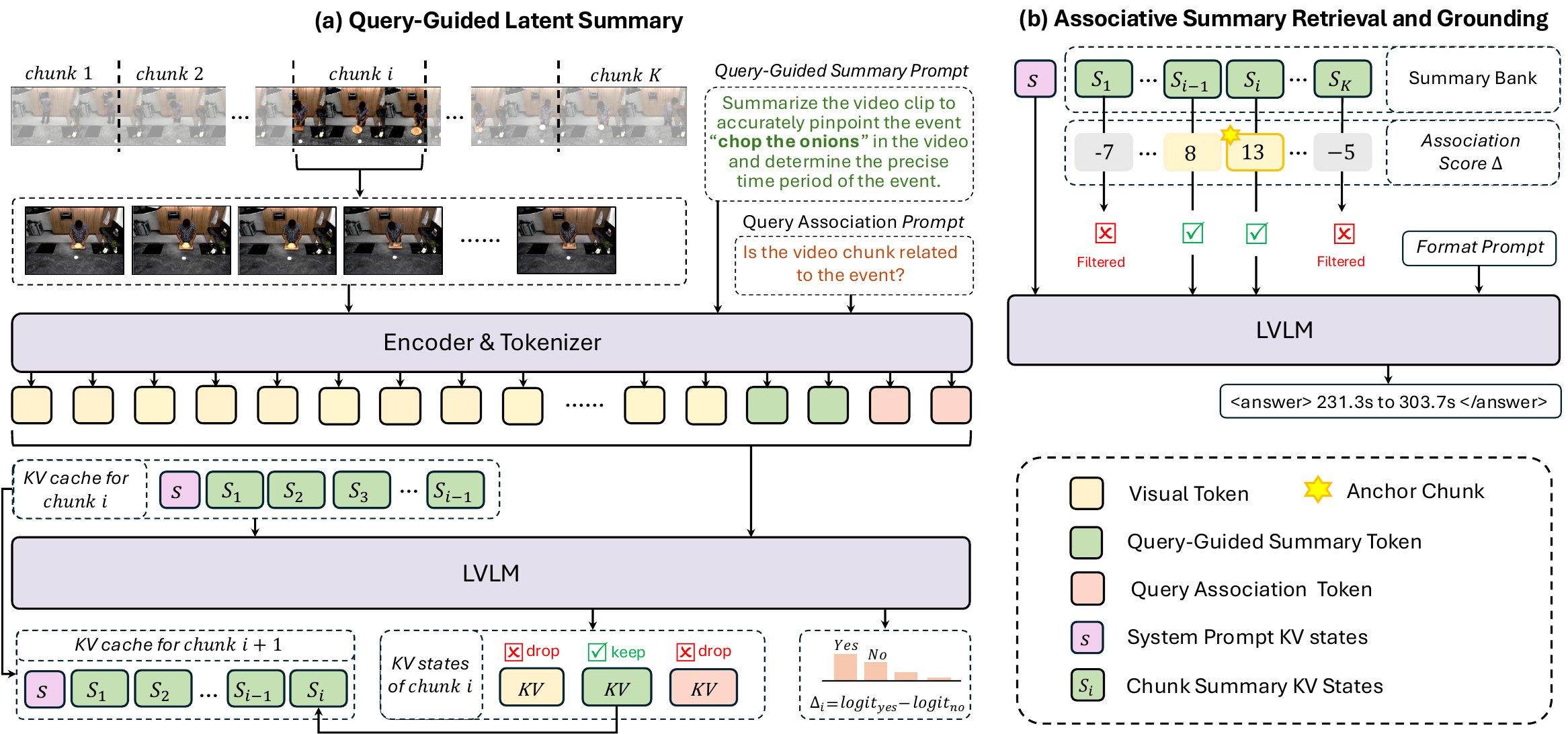}
    \caption{
    Framework of SumGround.
    (a) For each video chunk $C_i$, the LVLM jointly writes a query-guided summary state $S_i$ and estimates its query-association score $\Delta_i$, conditioned on previously retained summaries.
    Only the system and summary KV states are propagated to subsequent chunks, while dense visual states and the states used for association estimation are discarded.
    (b) After all chunks are processed, SumGround selects an anchor and retrieves a contiguous set of query-associated summaries according to their association scores.
    The retrieved summaries are used to generate the final temporal grounding boundaries.
    }
    \label{fig:pipeline}
\end{figure*}

\section{Related Work}

\noindent \textbf{Temporal Grounding with LVLMs.}
Video temporal grounding aims to localize the temporal span corresponding to a natural-language query~\cite{gao2017tall,wang2024hawkeye}.
Earlier methods typically combine pretrained video--language representations~\cite{radford2021learning,devlin2019bert} with dedicated temporal localization architectures, including proposal-based~\cite{zhang2020learning}, regression-based~\cite{yuan2019find}, DETR-style~\cite{lei2021detecting}, and unified clip-level prediction models~\cite{lin2023univtg}.
Recent LVLM-based methods instead formulate grounding as generative timestamp prediction and improve temporal awareness through normalized timestamp prediction~\cite{huang2024vtimellm,wang2024hawkeye}, temporal cues injected into visual tokens~\cite{ren2024timechat,guo2025vtg}, or explicit textual timestamps~\cite{yuan2025date}.
These methods primarily improve temporal representation and timestamp decoding, but pay less attention to how query-relevant visual evidence can be identified from long videos and organized into a focused context for precise boundary prediction.

\noindent \textbf{Long-Video Processing and Temporal Evidence Selection.}
Scaling LVLMs to long videos is challenging because dense visual tokens quickly exhaust practical context and memory budgets.
Dedicated long-video LVLMs improve efficiency through token-efficient encoders~\cite{li2024llama,ren2024timechat}, memory or summarization tokens~\cite{song2024moviechat,shu2025video}, and spatiotemporal compression~\cite{shen2024longvu}.
Related long-context and retrieval-based methods further reduce the effective context through chunk compression or selective retrieval~\cite{chevalier2023adapting,an2025lcirc,lin2025refrag}.
However, these capabilities often require dedicated modules or substantial additional training, and are not explicitly designed to retain the fine-grained evidence needed for precise temporal boundaries.

Recent long-video temporal grounding methods built on general-purpose LVLMs adopt grounded instruction tuning~\cite{zeng2024timesuite}, SFT-based coarse-to-fine localization~\cite{li2025universal}, or agentic localization workflows~\cite{liu2025videomind}.
Nevertheless, task-specific tuning may weaken the broader capabilities of the base LVLM, while coarse-to-fine localization remains constrained by its initial proposals.
These limitations motivate accumulating and selecting query-relevant evidence across the full video before predicting the final temporal boundaries.

\noindent \textbf{RLVR for Temporal Grounding.}
Recent RLVR-based methods optimize LVLMs for temporal grounding with verifiable rewards derived from temporal overlap~\cite{wang2025time,zhang2025timelens}.
This paradigm directly aligns timestamp generation with grounding quality and can better preserve the general capabilities of LVLMs~\cite{wang2025time}.
Nevertheless, existing RLVR methods typically optimize timestamp generation from a fixed, sparsely sampled visual context. They leave open how query-relevant evidence can be accumulated from dense long-video chunks while keeping RLVR optimization memory-efficient.

\section{Method}
\label{sec:method}
Given a video and a language query $q$, VTG aims to predict the temporal interval $[\hat{t}_s,\hat{t}_e]$ corresponding to the queried event.
An LVLM-based temporal grounding process can be viewed as a multimodal prefill to gather information, followed by autoregressive decoding to generate the answer.
As illustrated in Figure~\ref{fig:pipeline}, SumGround replaces a single full-video prefill with sequential video chunk prefills that retain compact query-guided latent summaries and record query-association scores. 
After all video chunks are processed, the association scores are used to retrieve a focused summary context for final decoding.
Length-aware gradient gating further enables RLVR training on long videos by selectively blocking visual-token gradients based on video length.

\subsection{SumGround Overview}
\label{sec:chunkwise_processing}
Given a video of duration $T$, we divide it into $K$ consecutive video chunks $\{C_i\}_{i=1}^{K}$.  
Each chunk $C_i$ contains the frames within the temporal interval $[\tau_{i-1},\tau_i)$, where $i\in\{1,...,K\}, \tau_0 =0, \tau_K=T$. 
Following~\cite{yuan2025date}, we insert the corresponding timestamp text after each frame. 

Let $\mathcal{F}_{\theta}^{\mathrm{prefill}}$ and $\mathcal{F}_{\theta}^{\mathrm{gen}}$ denote the prefill and autoregressive decoding stages of the LVLM parameterized by $\theta$, respectively.
For the $i$-th prefill pass, let $X_i$ denote the new input constructed for chunk $C_i$, and let $\mathcal{H}_{i-1}$ denote the KV cache retained from preceding passes.
The chunk-wise prefill is 
\begin{equation}
\left(
\mathrm{KV}_i,
\mathbf{z}_i
\right)
=
\mathcal{F}_{\theta}^{\mathrm{prefill}}
\left(
\mathcal{H}_{i-1},
X_i
\right),
\end{equation}
where $\mathrm{KV}_i$ denotes the newly computed KV states of $X_i$ and $\mathbf{z}_i$ denotes the output logits of the current pass.
After each prefill, SumGround retains the query-guided summary states from $\mathrm{KV}_i$, and derives the query-association score $\Delta_i$ from $\mathbf{z}_i$.
The retained query-guided summary states are used to update the KV cache from $\mathcal{H}_{i-1}$ to $\mathcal{H}_i$ which is passed to the prefill pass for the next chunk $i+1$. 
The association scores are collected across chunks for later retrieval.
No autoregressive decoding is performed until all chunk prefills are completed.

After all $K$ chunks have been processed, the association scores are used to retrieve a concentrated grounding context $\mathcal{H}^{*}$.
The temporal boundaries are then decoded as
\begin{equation}
(\hat{t}_s,\hat{t}_e)
=
\mathcal{F}_{\theta}^{\mathrm{gen}}
\left(
\mathcal{H}^{*},p_{\mathrm{fmt}}
\right).
\label{eq-decode}
\end{equation}
where $p_{\mathrm{fmt}}$ is the output format prompt.

The following subsections describe how query-guided summary states are constructed and exploited, how association scores are estimated, and how estimated association scores guide the retrieval of the final grounding context $\mathcal{H}^{*}$.


\subsection{Query-Guided Latent Summary}
\label{sec:sequential_summarization}

\noindent \textbf{Query-Guided Prompt KV states as latent summary.}
Recent analyses~\citep{zhang2025cross,yin2025lifting} show that visual evidence is progressively integrated into the contextualized KV states of prompt tokens. 
During subsequent generation, later tokens then rely primarily on these prompt KV states, rather than relying on the original visual-token states.
This suggests that KV states of prompt tokens can naturally serve as a compact latent representation of visual evidence.
Based on this observation, we append a query-guided summary prompt $p_{\mathrm{sum}}(q)$ 
after each video chunk,
\emph{e.g.},
\textcolor{black}{
\textit{``Summarize the video clip to accurately pinpoint the event
`chop the onions' and determine its precise temporal interval.''}, where \textit{`chop the onions'} is the content of query}.
The prompt asks the LVLM to summarize the visual evidence relevant to the queried event.
Without requiring time-consuming autoregressive textual summary generation or an additional summarization module, we directly retain the KV states of $p_{\mathrm{sum}}(q)$ as latent summaries:
\begin{equation}
S_i
=
\mathrm{KV}_i
\left[
p_{\mathrm{sum}}(q)
\right],
\end{equation}
Although the prompt text is shared across chunks, its contextualized KV states depend on the current visual tokens and the KV states of previous summaries, allowing each $S_i$ to encode different evidence. The ablation results in Table~\ref{tab:summary_ablation} further verify this design.

\noindent \textbf{Chunk-wise summary construction.}
Let $p_s$ denote the system prompt and $p_{\mathrm{assoc}}$ denote the association prompt detailed in the next subsection.
The new input for each prefill pass is
\begin{equation}
X_i
=
\begin{cases}
[p_s,C_1,p_{\mathrm{sum}}(q),p_{\mathrm{assoc}}],
& i=1,\\
[C_i,p_{\mathrm{sum}}(q),p_{\mathrm{assoc}}],
& i>1.
\end{cases}
\end{equation}
The first pass is performed without a preceding cache, whereas each subsequent input $X_i$ is prefilled conditioned on the cache $\mathcal{H}_{i-1}$ retained from earlier passes.

From the resulting KV states, we retain the system-prompt states
$s=\mathrm{KV}_1[p_s]$ from the first pass and the summary states
$S_i=\mathrm{KV}_i[p_{\mathrm{sum}}(q)]$ from each pass.
After processing vision chunk $C_i$, the cache becomes
\begin{equation}
    \mathcal{H}_i
    =
    \{s,S_1,\ldots,S_i\}.
\end{equation}
Thus, only the system and query-guided summary states are forwarded across chunks, while the  visual states are discarded.
After all $K$ chunks have been processed, SumGround obtains the summary bank $\{S_i\}_{i=1}^{K}$, storing compact query-relevant evidence aggregated from the whole video.



\subsection{Associative Summary Retrieval and Grounding}
\label{sec:associative_retrieval}

\noindent \textbf{Query-association estimation.}
Although the summaries are query-guided, summaries from less related chunks may still mislead the grounding. 
Rather than introducing a separate retriever or repeatedly invoking the model to refine the temporal interval, we estimate the association between chunk memory and query within the same prefill pass.
Specifically, the query-association prompt $p_{\mathrm{assoc}}$ asks whether the current chunk contains evidence associated with the queried event.
Because $p_{\mathrm{assoc}}$ follows $p_{\mathrm{sum}}(q)$ under causal attention, its computation does not affect the summary states $S_i$.

Let $\mathbf{z}_i$ denote the next-token prediction logits of the last token of $p_{\mathrm{assoc}}$.
We define the query association score as the logit margin between \texttt{Yes} and \texttt{No}:
\begin{equation}
\Delta_i
=
[\mathbf{z}_i]_{v_{\mathrm{yes}}}
-
[\mathbf{z}_i]_{v_{\mathrm{no}}},
\end{equation}
where $v_{\mathrm{yes}}$ and $v_{\mathrm{no}}$ are the vocabulary indices of \texttt{Yes} and \texttt{No}, respectively.
After each prefill, the KV states of $p_{\mathrm{assoc}}$ are discarded, while $\Delta_i$ is recorded for retrieval.

\noindent \textbf{Associative summary retrieval.}
We normalize the association scores across all chunks to form an anchor-selection distribution:
\begin{equation}
w_i
=
\frac{\exp(\Delta_i)}
{\sum_{j=1}^{K}\exp(\Delta_j)}.
\end{equation}
During inference, the anchor is selected as
\begin{equation}
a
=
\arg\max_i w_i,
\end{equation}
whereas during RLVR training it is sampled as
\begin{equation}
a
\sim
\operatorname{Categorical}(w_1,\ldots,w_K).
\end{equation}
The anchor identifies the chunk most likely to contain the queried event, while its full temporal extent may span adjacent chunks.
We expand the retrieved interval from the anchor in both directions, including consecutive chunks with positive association scores. 
Let $\mathcal{R}^{+}$ denote the maximal contiguous chunk index set, which contains $a$ and satisfies $\Delta_i>0, \forall i \in \mathcal{R}^{+}$; if $\Delta_a\leq0$, we set $\mathcal{R}^{+}=\varnothing$.
We set a minimum retrieval size $M$, where $M \leq K$.
If $|\mathcal{R}^{+}|\geq M$, we set $\mathcal{R}=\mathcal{R}^{+}$.
Otherwise, we retrieve an anchor-centered window to form $\mathcal{R}$ with $M$ consecutive chunks.
Finally, only the summaries indexed by $\mathcal{R}$ are exposed to final grounding.


\noindent \textbf{Grounding with the retrieved summaries.}
Let $S_{\mathcal{R}}$ denote the retrieved summaries.
When $|\mathcal{R}^{+}|\leq 1$, the temporal interval of queried event is likely to be small, and thus finer-grained cues may be required for precise boundary prediction. 
We therefore additionally include the visual cache $\mathcal{C}^{*}$ of the chunk with the highest \textcolor{black}{association score} to provide finer-grained evidence; otherwise,
$\mathcal{C}^{*}=\varnothing$.
During chunk-wise processing, we maintain the visual KV cache of the highest-scoring chunk seen so far, 
making $\mathcal{C}^{*}$ available for final grounding without extra forward passes.
The retrieved concentrated grounding context is constructed as
\begin{equation}
\mathcal{H}^{*}
=
\{s,S_{\mathcal{R}},\mathcal{C}^{*}\}.
\end{equation}
It is then passed together with the output format prompt $p_{\mathrm{fmt}}$ (see Eq.~(\ref{eq-decode})) to predict the temporal grounding boundaries.

\subsection{Length-Aware Gradient Gating for RLVR Training}
\label{sec:training}

\noindent \textbf{Length-aware gradient gating.}
Although we only retain compact summary states across chunk-wise prefills, back-propagating gradients through the dense visual tokens of a long video may still incur a high memory load during training. 
Let $N$ denote the number of frames in a training video, and $W$ be the maximum length where visual-token gradients can be retained under the available memory budget.

We define the length-aware gradient gate as
\begin{equation}
\gamma
=
\mathbb{I}[N\leq W],
\end{equation}
where $\mathbb{I}[\cdot]$ is the indicator function, \emph{i.e.,} when $N\le W$, $\gamma =1$, otherwise, $\gamma=0$.

Let $Z_i$ denote the visual-token states of chunk $C_i$.
During training, we replace $Z_i$ with $\widetilde{Z}_i$, which is
\begin{equation}
\widetilde{Z}_i
=
\gamma Z_i
+
(1-\gamma)\operatorname{sg}(Z_i),
\end{equation}
where $\operatorname{sg}(\cdot)$ denotes stop-gradient. 

As a result, for video samples with $N\leq W$, gradients from the RLVR objective backpropagate through the visual-token states, enabling the model to learn how query-relevant visual evidence is extracted and encoded into the retained summary states.
For samples with $N>W$, visual tokens are still used for forward summary construction but detached during backpropagation, while summary accumulation, retrieval, and grounding remain optimized.

\noindent \textbf{RLVR objective.}
We instantiate RLVR with Group Relative Policy Optimization (GRPO).
For each training instance, the current policy generates $G$ rollouts.
In rollout $g$, an anchor $a^g$ is sampled from the query-association distribution, the corresponding summary context is retrieved, and the model predicts a temporal interval $\hat{I}^g=[\hat{t}_s^g,\hat{t}_e^g]$.

Let $I^\star=[t_s,t_e]$ denote the ground-truth interval.
We define a soft association target for each chunk according to its temporal overlap with $I^\star$:
\begin{equation}
y_i
=
\frac{
|[\tau_{i-1},\tau_i)\cap I^\star|
}{
|I^\star|
},
\end{equation}
where $|\cdot|$ denotes the temporal duration.
All $G$ rollouts contribute to association optimization.
For grounding, we retain only rollouts whose retrieved context contains at least one chunk with $y_i>0$.

Each rollout receives separate rewards for anchor selection and temporal grounding:
\begin{equation}
r_{\mathrm{assoc}}^g
=
y_{a^g},
\qquad
r_{\mathrm{tg}}^g
=
\operatorname{IoU}(\hat{I}^g,I^\star).
\end{equation}

The two rewards evaluate different actions, so we normalize their group-relative advantages separately and yield the GRPO objectives $\mathcal{J}_{\mathrm{assoc}}$ and $\mathcal{J}_{\mathrm{tg}}$, respectively.
The final objective is
\begin{equation}
\theta^\star
= \arg\max_{\theta} \left(
\mathcal{J}_{\mathrm{tg}} + \lambda\mathcal{J}_{\mathrm{assoc}} \right), 
\end{equation}
where $\lambda$ balances the two objectives.
Details of advantage normalization, clipped policy optimization, and reference-policy KL regularization are provided in the supplementary material.

\begin{table*}[t]
\centering
\scriptsize
\setlength{\tabcolsep}{3.0pt}
\renewcommand{\arraystretch}{1.05}
\resizebox{\textwidth}{!}{
\begin{tabular}{lcccccccccccc}
\toprule
\multirow{3}{*}{\textbf{Method}}
& \multicolumn{6}{c}{\textbf{Long-video Benchmarks}}
& \multicolumn{6}{c}{\textbf{Standard-video Benchmarks}} \\
\cmidrule(lr){2-7}
\cmidrule(lr){8-13}

& \multicolumn{3}{c}{\textbf{Ego4D-NLQ}}
& \multicolumn{3}{c}{\textbf{TACoS}}
& \multicolumn{3}{c}{\textbf{Charades-STA}}
& \multicolumn{3}{c}{\textbf{ActivityNet-Captions}} \\
\cmidrule(lr){2-4}
\cmidrule(lr){5-7}
\cmidrule(lr){8-10}
\cmidrule(lr){11-13}

& R1@.3 & R1@.5 & mIoU 
& R1@.3 & R1@.5 & mIoU 
& R1@.5 & R1@.7 & mIoU 
& R1@.5 & R1@.7 & mIoU \\
\midrule

\multicolumn{13}{l}{\textit{Supervised / SFT-based grounding methods}} \\
\midrule

UniVTG~\cite{lin2023univtg} 
& 6.48 & 3.48 & 4.63 
& 5.17 & 1.27 & 4.40 
& 25.22 & 10.03 & 27.12 
& 11.10 & 4.06 & 16.86 \\

VTG-LLM~\cite{guo2025vtg} 
& 1.71 & 0.46 & 1.36 
& 6.87 & 2.92 & 5.27 
& 34.11 & 15.81 & 34.93 
& 12.32 & 6.74 & 17.86 \\

TimeSuite~\cite{zeng2024timesuite} 
& 0.88 & 0.43 & 0.94 
& 6.75 & 2.50 & 5.71 
& 48.95 & 24.65 & 45.91 
& 16.56 & 9.28 & 22.03 \\

VideoMind~\cite{liu2025videomind}
& 7.20 & 3.70 & 5.40
& 49.50 & \underline{36.20} & \underline{34.40}
& 59.10 & 31.20 & 50.20
& 30.30 & 15.70 & 33.30 \\

UniTime~\cite{li2025universal}
& \underline{14.67} & 7.38 & \underline{10.18} 
& \underline{50.06} & 31.54 & 33.38 
& 59.09 & 31.88 & 52.19 
& 22.77 & 14.14 & 27.31 \\

\midrule
\multicolumn{13}{l}{\textit{RLVR-based grounding methods with released checkpoints}} \\
\midrule

Time-R1~\cite{wang2025time}
& \cellcolor{gray!20} 4.82 & \cellcolor{gray!20} 2.51 & \cellcolor{gray!20} 3.51
& \cellcolor{gray!20} 30.54 & \cellcolor{gray!20} 17.72 & \cellcolor{gray!20} 20.54
& 60.80 & 35.30 & \textbf{58.10}
& \textbf{39.00} & \textbf{21.40} & \textbf{40.50} \\

TimeLens~\cite{zhang2025timelens}
& \cellcolor{gray!20} 13.48 & \cellcolor{gray!20} \underline{7.56} & \cellcolor{gray!20} 9.27
& \cellcolor{gray!20} 47.40 & \cellcolor{gray!20} 35.20 & \cellcolor{gray!20} 32.77
& 39.80 & 14.50 & 42.30
& 35.20 & 19.70 & 37.30 \\

\midrule
\multicolumn{13}{l}{\textit{RLVR-based grounding methods retrained under the same training data}} \\
\midrule

Time-R1$^\dagger$~\cite{wang2025time}
& \cellcolor{gray!20} 1.95 & \cellcolor{gray!20} 0.63 & \cellcolor{gray!20} 1.68
& \cellcolor{gray!20} 21.49 & \cellcolor{gray!20} 11.05 & \cellcolor{gray!20} 15.20
& \cellcolor{gray!20} 57.74 & \cellcolor{gray!20} 31.53 & \cellcolor{gray!20} 51.19
& \cellcolor{gray!20} 26.39 & \cellcolor{gray!20} 12.39 & \cellcolor{gray!20} 29.99 \\

TimeLens$^\dagger$~\cite{zhang2025timelens}
& \cellcolor{gray!20} 10.66 & \cellcolor{gray!20} 5.37 & \cellcolor{gray!20} 7.87
& \cellcolor{gray!20} 44.80 & \cellcolor{gray!20} 28.11 & \cellcolor{gray!20} 29.90
& \cellcolor{gray!20} \textbf{66.48} & \cellcolor{gray!20} 34.38 & \cellcolor{gray!20} 54.85
& \cellcolor{gray!20} 36.03 & \cellcolor{gray!20} 18.11 & \cellcolor{gray!20} 38.43 \\

\midrule
\multicolumn{13}{l}{\textit{Ours}} \\
\midrule

SumGround w/o. Retrieval
& 4.81 & 1.98 & 4.06
& 33.79 & 19.77 & 23.70
& 65.35 & \underline{39.49} & 55.94
& 35.92 & 19.12 & 39.09 \\

SumGround
& \textbf{16.42} & \textbf{9.51} & \textbf{11.83}
& \textbf{52.96} & \textbf{38.07} & \textbf{35.93}
& \underline{65.91} & \textbf{39.57} & \underline{56.10}
& \underline{36.70} & \underline{20.72} & \underline{39.54} \\

\bottomrule
\end{tabular}
}
\caption{
Comparison with previous methods on four held-out benchmarks.
Ego4D-NLQ and TACoS are long-video benchmarks with average durations of 500s and 368s, respectively, while Charades-STA and ActivityNet-Captions with average durations of 29s and 118s.
$^\dagger$ denotes methods retrained with the same training data and evaluation pipeline as SumGround.
Gray-shaded entries denote results reproduced or evaluated by us. We use officially released checkpoints and codes.
Other numbers are directly cited from previous papers.
\emph{SumGround w/o. Retrieval} means we remove the associative summary retrieval operation from our method.
Best and second-best results are shown in \textbf{bold} and \underline{underlined}, respectively.
}
\label{tab:main_vtg}
\end{table*}


\begin{table}[t]
\centering
\resizebox{\columnwidth}{!}{
\begin{tabular}{lcccc}
\toprule
\textbf{Setting}
& \textbf{R1@0.3}
& \textbf{R1@0.5}
& \textbf{R1@0.7}
& \textbf{mIoU} \\
\midrule

\multicolumn{5}{l}{\textit{(a) Summary Type}} \\

\midrule

w/o. Summary
& 78.09 & 63.01 & 36.42 & 54.46 \\

\texttt{<summary>}
& 73.12 & 54.38 & 27.28 & 48.78 \\

Query-Agnostic
& 76.96 & 60.56 & 34.81 & 52.93 \\

Query-Guided (Ours)
& \textbf{81.83}
& \textbf{69.57}
& \textbf{44.30}
& \textbf{58.75} \\

\midrule

\multicolumn{5}{l}{\textit{(b) Chunk-wise Dependency}} \\

\midrule

Independent
& 72.39 & 56.02 & 30.38 & 49.53 \\

Causal (Ours)
& \textbf{82.04}
& \textbf{72.04}
& \textbf{48.25}
& \textbf{59.88} \\

\bottomrule
\end{tabular}
}
\caption{
Ablation of summary design on Charades-STA. 
All variants use approximately the same retained token budget. 
(a) Summary representation comparison 
The ``w/o. Summary'' directly grounds from the visual inputs without any summary operations.
``\texttt{<summary>}'' means learned \texttt{<summary>} tokens.
``Query-Agnostic'' means we use a general summary prompt which does not rely on the query content.
To simplify our comparison, we treat the whole video as a single chunk. 
(b) Chunk-wise Dependency comparison. ``Independent'' means the summary of each chunk is independent on those of other chunks. 
}
\label{tab:summary_ablation}
\end{table}

\begin{table}[t]
\centering
\small
\setlength{\tabcolsep}{5.5pt}
\renewcommand{\arraystretch}{1.05}
\resizebox{\columnwidth}{!}{
\begin{tabular}{lcccc}
\toprule
\textbf{Retrieval Strategy}
& \textbf{R1@0.3}
& \textbf{R1@0.5}
& \textbf{R1@0.7}
& \textbf{mIoU} \\
\midrule


Random
& 28.32
& 19.57
& 9.75
& 19.46 \\

Anchor Only
& 43.64
& 30.17
& 15.20
& 29.67 \\

Ours
& \textbf{52.96}
& \textbf{38.07}
& \textbf{18.67}
& \textbf{35.93} \\

\bottomrule
\end{tabular}
}
\caption{
Comparison of summary retrieval strategies on TACoS.
\emph{Random} retains the same number of summaries as our method but selects them at random.
\emph{Anchor Only} retains only the anchor summary.
}
\label{tab:retrieval_ablation}
\end{table}









\begin{table}[t]
\centering
\small
\setlength{\tabcolsep}{4.5pt}
\renewcommand{\arraystretch}{1.08}
\resizebox{\columnwidth}{!}{
\begin{tabular}{ccccc}
\toprule
\boldmath$\mathbf{\gamma}$
& \textbf{Long Video}
& \textbf{R1@0.3}
& \textbf{R1@0.5}
& \textbf{mIoU} \\
\midrule

$1$
& \checkmark
& OOM
& OOM
& OOM \\

$0$
& \checkmark
& 27.44
& 14.32
& 20.42 \\

$1$
& \xmark
& 45.66
& 29.94
& 31.46 \\

$\mathbb{I}[N\leq W]$
& \checkmark
& \textbf{52.96}
& \textbf{38.07}
& \textbf{35.93} \\

\bottomrule
\end{tabular}
}
\caption{
Comparison of gradient-gating configurations on TACoS. $\gamma$ controls whether gradients propagate through visual-token states ($\gamma$=1 and 0 means retaining and stopping visual gradients respectively), and \emph{Long Video} means long videos with $N>W$ are utilized in training. The total number of training samples is kept the same across all variants. The last row denotes length-aware gradient gating. \emph{OOM} denotes out of memory.
}
\label{tab:lgg_ablation}
\end{table}


\begin{figure}[t]
    \centering
    \includegraphics[width=\columnwidth]{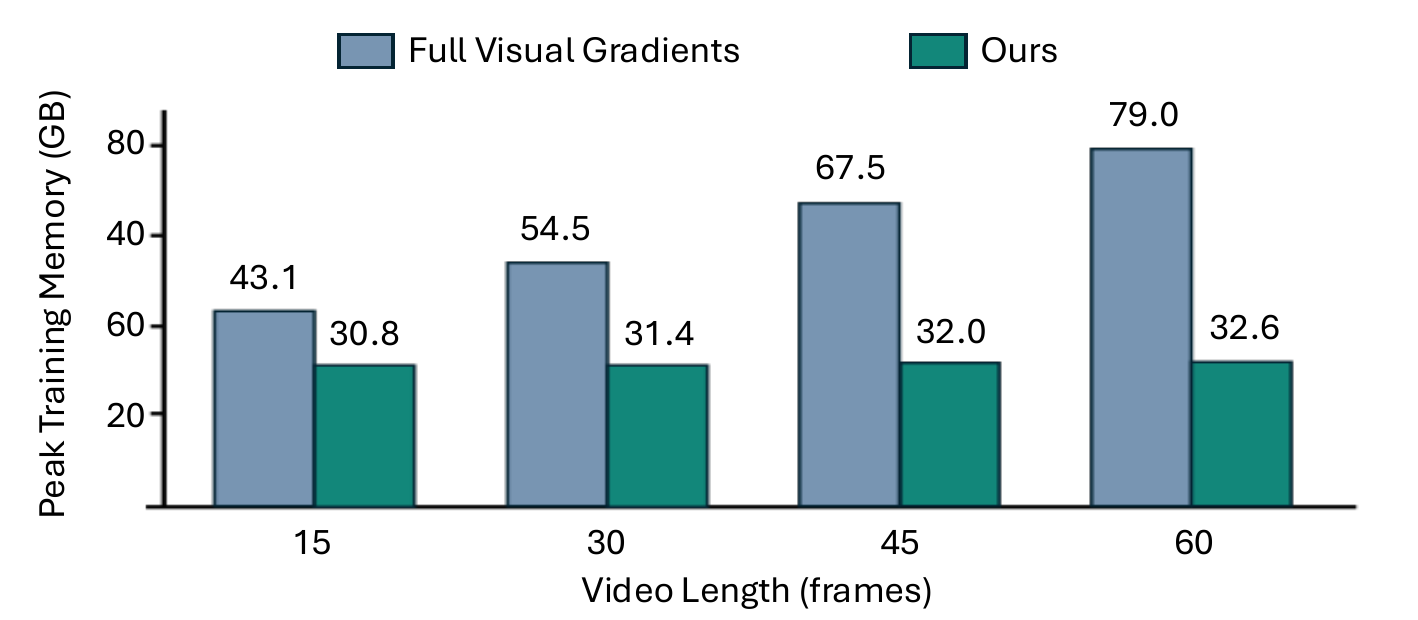}
    \caption{
        Peak GPU memory during training as the video length increases from 15 to 60 frames.
        All inputs are divided into 15-frame chunks.
        \emph{Full Visual Gradients} retains gradients through all visual tokens, while \emph{ours} applies length-aware gradient gating technique.
    }
    \label{fig:memory_scaling}
\end{figure}

\section{Experiments}

\subsection{Setup}

\noindent \textbf{Datasets.}
We construct the training set by randomly sampling 42.5K examples from datasets including
NaQ~\cite{ramakrishnan2023naq}, 
DiDeMo~\cite{anne2017localizing}, QuerYD~\cite{oncescu2021queryd}, 
HiRest~\cite{zala2023hierarchical}, COIN~\cite{tang2019coin}, 
Momentor~\cite{qian2024momentor}, and YouCook2~\cite{zhou2018towards}. 
For video temporal grounding evaluation, we use four standard benchmarks.
For short-video temporal grounding, we evaluate on Charades-STA~\citep{gao2017tall}, which focuses on indoor human activities, and ActivityNet-Captions~\citep{krishna2017dense}, which contains diverse videos with dense event annotations. For long-video temporal grounding, we use TACoS~\citep{regneri2013grounding}, mainly consisting of cooking videos, and Ego4D-NLQ~\citep{grauman2022ego4d}, which pairs long egocentric videos with natural-language queries.
We evaluate a single checkpoint across all four benchmarks in a zero-shot setting, without dataset-specific fine-tuning.
Detailed dataset descriptions are provided in the supplementary material.

\noindent \textbf{Evaluation Metrics.}
Following~\cite{li2025universal}, we report Recall@1 (R1@) at Intersection-over-Union (IoU) thresholds and mean IoU (mIoU) for evaluation.
Specifically, we use IoU thresholds of 0.3 (\emph{i.e.,} R1@.3) and 0.5 for long-video benchmarks, and 0.5 and 0.7 for short-video benchmarks.


\noindent \textbf{Implementation Details.}
We use Qwen2.5-VL-7B-Instruct~\cite{bai2025qwen2} as the backbone LVLM and freeze the vision encoder during training. 
The minimum retrieval size $M$ is set to $3$.
For GRPO optimization, each training instance uses $G=8$ rollouts. 
The objective weight is set to $\lambda=0.1$. 
We set the gradient-gating threshold $W$ to 42 frames and use  chunk size of 42 frames (except for the last chunk, which may contain fewer than 42 frames) during inference. 
For short video samples with $N \leq W$, we randomly choose the number of chunks $K\in\{1,2,3\}$.
For long video samples with $N>W$, we partition videos with each chunk of $W$ frames.
We train the model with DeepSpeed ZeRO-3 using a learning rate of $1\times10^{-6}$ for 2 epochs on 8 NVIDIA H20 GPUs.

\subsection{Comparison with Previous State-of-the-Arts}
\label{sec:sota_comparison}

Table~\ref{tab:main_vtg} compares SumGround with previous state-of-the-art (SOTA) methods on four typical video benchmarks, including long-video benchmarks (\emph{i.e.,} Ego4D-NLQ and TACoS) and short-video benchmarks (\emph{i.e.,} Charades-STA and ActivityNet-Captions).
Note that we evaluate SumGround in a zero-shot setting, without fine-tuning on each evaluation benchmark.
Previous methods can be roughly categorized into two groups, including supervised/SFT-based methods (\emph{e.g.,} UniVTG, VTG-LLM, \emph{etc.} ) and RLVR-based methods (\emph{e.g.,} Time-R1, TimeLens).
Our method falls into the category of RLVR-based methods. 
For previous RLVR-based methods (\emph{i.e.,} Time-R1, TimeLens), besides the released checkpoints, we also retrain and evaluate the models with their official released codes under our training data to enable a fair comparison.

From Table~\ref{tab:main_vtg}, we have three observations.
First, generally speaking, the performance on long-video benchmarks is obviously lower than that on short-video benchmarks, either for SFT-based or for RLVR-based methods.
This indicates that it is more challenging to locate sparse evidence in long visual contexts.
Second, comparing previous RLVR-based methods to SFT-based methods, we observe that previous RLVR-based methods perform favorably against SFT-based methods on short-video benchmarks, but generally performs worse than SOTA SFT-based methods on long-video benchmarks.
This suggests that without specific design, purely RLVR cannot effectively mine and exploit query-relevant evidence for grounding.
Third, our SumGround performs favorably against all previous works on both short-video and long-video benchmarks.
Note that under the same training data, our method performs obviously better previous RLVR-based methods. For example, on long-video benchmarks Ego4D-NLQ and TACoS, our method outperforms TimeLens$^\dagger$ by around 6\% and 8\% with R1@.3 and 4\% and 6\% with mIoU.
Overall, the comparisons verify the superiority of our method in performing long-video temporal grounding, which benefits from our specific design to enable LVLM to effectively mine and exploit query-relevant evidence from long videos.



\section{Ablation Study}

\noindent \textbf{Effect of summary representation.}
Table~\ref{tab:summary_ablation}(a) examines how the summary representation affects grounding. To simplify our experiment and guarantee a fair comparison, we consider single chunk and keep summary length (tokens) approximately the same.
We compare direct visual grounding (``w/o. Summary'') with three summary variants: learned \texttt{<summary>} tokens, a query-agnostic natural-language summary prompt, and our query-guided summary prompt.
Only our query-guided design can surpass direct visual grounding. 
These results show latent compression alone is insufficient and our query-guided way can effectively extract query-relevant information to support grounding.


\noindent \textbf{Effect of causal accumulation.}
Table~\ref{tab:summary_ablation}(b) demonstrates how chunk-wise dependency affects the performance. 
Comparing with the ``Independent'' way where the summary of each chunk is independent of other chunks, 
the causal way (ours) performs remarkably better. 
The results verify that the summary of current chunk should attend to the summaries of previous chunks for better information aggregation.


\noindent \textbf{Effect of associative summary retrieval.}
Table~\ref{tab:main_vtg} shows that without associative summary retrieval, the performance obviously drops, especially for long videos.
Further, in Table~\ref{tab:retrieval_ablation}, we compare our retrieval strategy with several variants to verify the effectiveness of our design.
Among all variants, our method performs the best.
All the results verify that associative summary retrieval may effectively remove redundant summaries and ease the grounding.

\noindent \textbf{Effect of RLVR on associative summary retrieval capability.}
To verify the effectiveness of RLVR on enhancing associative summary retrieval capability, we replace the learned association scores with those estimated by the base LVLM, while keeping the trained SumGround model for final grounding.
This variant only achieves 30.35 mIoU on TACoS, worse than ours (35.93 mIoU), showing that RLVR improves the query-association estimation and thus improves associative summary retrieval capability.


\noindent \textbf{Effect of length-aware gradient gating.}
Table~\ref{tab:lgg_ablation} compares our length-aware gradient gating with several variants.
Keeping visual gradients for all inputs ($\gamma=1$) is infeasible due to OOM, whereas stopping them ($\gamma=0$) for all inputs substantially degrades grounding, showing that visual gradients are necessary for learning effective summary.
Compared with training only on short contexts, length-aware gradient gating further improves mIoU by 4.47 points.
This result shows that short-context visual-to-summary learning and long-context summary optimization are complementary.


Figure~\ref{fig:memory_scaling} examines how peak GPU memory during training increases as video length increases.
We compare our method to the variant which enables full visual gradients.
As the length increases from 15 to 60 frames, the memory  of ``full visual gradients'' is increased by around 84\%. 
In contrast, our method only increases 6\%. 
Note that at 60 frames, our method consumes less than half the memory of that variant.
This comparison demonstrates the our memory efficiency. 

\section{Conclusion}
We propose a novel framework named SumGround to perform query-guided chunk condensation to aggregate and retrieve query-relevant evidence.
Specifically, we split the video into several chunks and perform two-level chunk condensation, including query-guided latent summary and associative summary retrieval.
Both capabilities are enabled by RLVR. 
To reduce memory consumption, we propose a length-aware gradient gating module to selectively stop gradient back-propagated to visual tokens.
Experiments verify the effectiveness of our method. 
We expect our attempts may inspire future investigations on better mining query-relevant evidence for effective long video temporal grounding.
%

\bibliography{aaai2027}

\section*{Appendix}

\section{Inference procedure.}
During inference, SumGround follows a deterministic summary-retrieval-and-grounding pipeline.
Given a video-query pair, it first samples timestamped frames and partitions the video into $K$ consecutive chunks.
Each chunk-wise pass jointly writes a query-guided summary state $S_i$ and estimates a query-association score $\Delta_i$.
Throughout chunk-wise processing, SumGround retains the system KV states $s$, accumulates the summary states $\{S_i\}_{i=1}^{K}$, and maintains the local visual KV cache of the highest-scoring chunk observed so far.
After all chunks are processed, it computes the query-association distribution
$p_i=\operatorname{softmax}(\Delta)_i$
and selects the anchor chunk as
$a=\arg\max_i p_i$.
Starting from $a$, SumGround constructs the retrieved chunk set $\mathcal{R}$ using the associative summary retrieval rule defined the main paper, and concatenates the corresponding summary states as $S_{\mathcal{R}}$.
When no non-anchor chunk has positive query association, 
the retained local visual cache is included as $\mathcal{C}^{*}$; otherwise, $\mathcal{C}^{*}=\varnothing$.
Finally, the model predicts the temporal interval conditioned on
$[s,S_{\mathcal{R}},\mathcal{C}^{*},p_{\mathrm{fmt}}]$.

\section{Training Objective and Procedure}
\label{app:training_details}

\subsection{Task-Specific GRPO Objectives}
\label{app:grpo_details}

We jointly optimize association estimation and temporal grounding
with task-specific GRPO objectives.
The two objectives differ in their action spaces, valid rollout sets,
and regularization.
Association optimization treats the sampled anchor chunk as a
categorical action, whereas temporal grounding optimizes the generated
response tokens.
We apply reference-policy KL regularization only to temporal grounding.

\paragraph{Task-specific rollout sets.}
Let
\begin{equation}
\mathcal{Y}^{+}
=
\{i \mid y_i>0\}
\end{equation}
denote the set of chunks that temporally overlap with the
ground-truth interval.

For rollout $g$, let $\mathcal{R}^{g}$ denote the indices of all video
chunks represented in its final grounding context, including the
retrieved summary states and, when used, the additional local visual
cache.
We define the task-specific rollout sets as
\begin{equation}
\mathcal{S}_{\mathrm{assoc}}
=
\{1,\ldots,G\},
\qquad
\mathcal{S}_{\mathrm{tg}}
=
\left\{
g
\;\middle|\;
\mathcal{Y}^{+}\cap\mathcal{R}^{g}\neq\emptyset
\right\}.
\label{eq:task_rollout_sets}
\end{equation}

Every sampled anchor defines a valid association action, so all
$G$ rollouts are included in $\mathcal{S}_{\mathrm{assoc}}$.
For temporal grounding, we retain only rollouts whose grounding
context contains ground-truth-relevant evidence.
This prevents the timestamp-generation policy from being penalized
for failures caused solely by an evidence-deficient retrieved context.

\paragraph{Task-specific advantage normalization.}
For each objective
$\kappa\in\{\mathrm{assoc},\mathrm{tg}\}$,
we compute the reward mean and standard deviation over its corresponding
rollout set:
\begin{equation}
\mu_{\kappa}
=
\frac{1}{|\mathcal{S}_{\kappa}|}
\sum_{g\in\mathcal{S}_{\kappa}}
r_{\kappa}^{g},
\end{equation}
\begin{equation}
\sigma_{\kappa}
=
\sqrt{
\frac{1}{|\mathcal{S}_{\kappa}|}
\sum_{g\in\mathcal{S}_{\kappa}}
\left(
r_{\kappa}^{g}-\mu_{\kappa}
\right)^2
}.
\end{equation}

The group-relative advantage is
\begin{equation}
\widehat{A}_{\kappa}^{g}
=
\frac{
r_{\kappa}^{g}-\mu_{\kappa}
}{
\sigma_{\kappa}+\epsilon_{\mathrm{adv}}
},
\qquad
g\in\mathcal{S}_{\kappa},
\label{eq:task_advantage}
\end{equation}
where $\epsilon_{\mathrm{adv}}>0$ is a numerical-stability constant.
The normalized advantages are treated as constants during policy
optimization.

When $\mathcal{S}_{\mathrm{tg}}=\emptyset$, the temporal-grounding
objective is omitted for that training instance.
When all rewards in a rollout set are identical, the corresponding
advantages are zero, and that objective produces no policy-gradient
update for the instance.

\paragraph{Association objective.}
For association optimization, the action in rollout $g$ is the sampled
anchor chunk $a^{g}$.
Its probability under the current policy is
\begin{equation}
p_{\theta}(a^{g})
=
w_{a^{g}},
\end{equation}
where $w_i$ is the association distribution defined in the main paper.

Let $\theta_{\mathrm{old}}$ denote the rollout policy.
We define the probability ratio
\begin{equation}
\rho_{\mathrm{assoc}}^{g}
=
\frac{
p_{\theta}(a^{g})
}{
p_{\theta_{\mathrm{old}}}(a^{g})
},
\end{equation}
and its clipped counterpart
\begin{equation}
\bar{\rho}_{\mathrm{assoc}}^{g}
=
\operatorname{clip}
\left(
\rho_{\mathrm{assoc}}^{g},
1-\epsilon_{\mathrm{clip}},
1+\epsilon_{\mathrm{clip}}
\right),
\end{equation}
where $\epsilon_{\mathrm{clip}}$ is the policy-ratio clipping threshold.

The association objective is
\begin{equation}
\mathcal{J}_{\mathrm{assoc}}
=
\frac{1}{|\mathcal{S}_{\mathrm{assoc}}|}
\sum_{g\in\mathcal{S}_{\mathrm{assoc}}}
\min
\left(
\rho_{\mathrm{assoc}}^{g}
\widehat{A}_{\mathrm{assoc}}^{g},
\bar{\rho}_{\mathrm{assoc}}^{g}
\widehat{A}_{\mathrm{assoc}}^{g}
\right).
\label{eq:assoc_grpo}
\end{equation}

We do not apply reference-policy KL regularization to this objective,
because the anchor is a task-specific categorical action over video
chunks rather than a language-generation action.

\paragraph{Temporal-grounding objective.}
For each rollout $g\in\mathcal{S}_{\mathrm{tg}}$, let
\begin{equation}
\mathbf{o}^{g}
=
(o_{1}^{g},\ldots,o_{T_g}^{g})
\end{equation}
denote the generated response containing the predicted temporal
boundaries.
Let
\begin{equation}
H^g=[s,S_{\mathcal{R}^g},\mathcal{C}^g]
\end{equation}
denote the retrieved grounding context of rollout $g$.
The token-level history is
\begin{equation}
h_t^g=(H^g,p_{\mathrm{fmt}},o_{<t}^g).
\end{equation}
denote the grounding context and generated prefix preceding token
$o_t^g$.

The token-level probability ratio is
\begin{equation}
\rho_{\mathrm{tg},t}^{g}
=
\frac{
\pi_{\theta}
\left(
o_t^g \mid h_t^g
\right)
}{
\pi_{\theta_{\mathrm{old}}}
\left(
o_t^g \mid h_t^g
\right)
},
\end{equation}
with the clipped ratio
\begin{equation}
\bar{\rho}_{\mathrm{tg},t}^{g}
=
\operatorname{clip}
\left(
\rho_{\mathrm{tg},t}^{g},
1-\epsilon_{\mathrm{clip}},
1+\epsilon_{\mathrm{clip}}
\right).
\end{equation}

Let $\pi_{\mathrm{ref}}$ denote the frozen reference policy.
For each generated token, we use the sampled-token KL estimator
\begin{equation}
d_{\mathrm{KL},t}^{g}
=
\exp
\left(
\ell_{\mathrm{ref},t}^{g}
-
\ell_{\theta,t}^{g}
\right)
-
\left(
\ell_{\mathrm{ref},t}^{g}
-
\ell_{\theta,t}^{g}
\right)
-1,
\label{eq:sampled_kl}
\end{equation}
where
\begin{equation}
\ell_{\theta,t}^{g}
=
\log
\pi_{\theta}
\left(
o_t^g\mid h_t^g
\right),
\qquad
\ell_{\mathrm{ref},t}^{g}
=
\log
\pi_{\mathrm{ref}}
\left(
o_t^g\mid h_t^g
\right).
\end{equation}

The temporal-grounding objective is
\begin{equation}
\begin{aligned}
\mathcal{J}_{\mathrm{tg}}
=
\frac{1}{|\mathcal{S}_{\mathrm{tg}}|}
\sum_{g\in\mathcal{S}_{\mathrm{tg}}}
\frac{1}{T_g}
\sum_{t=1}^{T_g}
\Big[
&
\min
\left(
\rho_{\mathrm{tg},t}^{g}
\widehat{A}_{\mathrm{tg}}^{g},
\bar{\rho}_{\mathrm{tg},t}^{g}
\widehat{A}_{\mathrm{tg}}^{g}
\right)
\\
&-
\beta
d_{\mathrm{KL},t}^{g}
\Big],
\end{aligned}
\label{eq:tg_grpo}
\end{equation}
where $\beta$ controls the strength of reference-policy KL
regularization.

The KL term is applied only to temporal grounding to prevent excessive
drift in language generation while allowing the task-specific
association distribution to adapt freely.

\paragraph{Final training objective.}
The complete objective is
\begin{equation}
\theta^{\star}
=
\arg\max_{\theta}
\left(
\mathcal{J}_{\mathrm{tg}}
+
\lambda
\mathcal{J}_{\mathrm{assoc}}
\right),
\label{eq:full_grpo_objective}
\end{equation}
where $\lambda$ balances temporal-grounding and association
optimization.

\subsection{Training Algorithm}
\label{app:training_algorithm}

Algorithm~\ref{alg:sumground_training} summarizes the training procedure of SumGround.
It integrates causal accumulation of query-guided summaries, associative summary retrieval, length-aware gradient gating, and dual-objective GRPO.
Each chunk $C_i$ contains sampled frames together with explicit timestamp tokens.
For a training instance with $N$ sampled frames, length-aware gradient gating applies stop-gradient only to the visual-token states when $N>W$, while leaving the timestamp and prompt-token states unchanged.
This operation preserves the forward computation while reducing the memory required for back-propagation through long visual contexts.
The detailed reward definitions and GRPO objectives are given above.

\begin{algorithm}[t]
\caption{Training Procedure of SumGround}
\label{alg:sumground_training}
\small
\begin{algorithmic}[1]

\REQUIRE Training instance $(V,q,I^\star)$, policy
$\pi_\theta$, rollout number $G$, and gradient threshold $W$.

\STATE Sample $N$ timestamped frames and partition them into
chunks $\{C_i\}_{i=1}^{K}$.
\STATE Initialize summary bank $\mathcal{B}\leftarrow\emptyset$,
$\Delta^\star\leftarrow-\infty$, and
$\mathcal{C}^\star\leftarrow\emptyset$.

\FOR{$i=1$ to $K$}
    \STATE Apply length-aware gradient gating to the visual-token
    states of $C_i$, yielding $\widetilde{C}_i$.
    \STATE Set
    $X_i=[p_s,\widetilde{C}_i,p_{\mathrm{sum}}(q),p_{\mathrm{assoc}}]$
    if $i=1$; otherwise set
    $X_i=[s,S_{<i},\widetilde{C}_i,
    p_{\mathrm{sum}}(q),p_{\mathrm{assoc}}]$.
    \STATE Prefill $X_i$; retain $s$ when $i=1$, retain
    $S_i=\mathrm{KV}(X_i)[p_{\mathrm{sum}}(q)]$, and append
    $S_i$ to $\mathcal{B}$.
    \STATE Compute
    $\Delta_i=\ell_{\mathrm{yes}}^{(i)}
    -\ell_{\mathrm{no}}^{(i)}$.
    \IF{$\Delta_i>\Delta^\star$}
        \STATE Retain the current local visual cache as
        $\mathcal{C}^\star$ and update $\Delta^\star$.
    \ENDIF
    \STATE Discard the remaining chunk-wise KV states.
\ENDFOR

\STATE Compute $w_i=\operatorname{softmax}(\Delta)_i$.

\FOR{$g=1$ to $G$}
    \STATE Sample anchor
    $a^g\sim\operatorname{Categorical}(w_1,\ldots,w_K)$.
    \STATE Construct $(\mathcal{R}^g,\mathcal{C}^g)$ using the
    associative retrieval and auxiliary-cache rules.
    \STATE Generate
    $\hat{I}^g\sim
    \pi_\theta(\cdot\mid
    [s,S_{\mathcal{R}^g},\mathcal{C}^g,p_{\mathrm{fmt}}])$.
    \STATE Compute $r_{\mathrm{assoc}}^g$ and $r_{\mathrm{tg}}^g$.
\ENDFOR

\STATE Form $\mathcal{S}_{\mathrm{assoc}}$ and
$\mathcal{S}_{\mathrm{tg}}$, and normalize their rewards.
\STATE Compute $\mathcal{J}_{\mathrm{assoc}}$ and
$\mathcal{J}_{\mathrm{tg}}$ as defined above.
\STATE Update $\pi_\theta$ by maximizing
$\mathcal{J}_{\mathrm{tg}}
+\lambda\mathcal{J}_{\mathrm{assoc}}$.

\end{algorithmic}
\end{algorithm}

\begin{table*}[t]
\centering
\caption{
Statistics of temporal grounding datasets used in our experiments.
``Video Len.'' and ``Moment Len.'' denote the average video duration and average annotated temporal moment duration, respectively.
}
\label{tab:dataset_statistics}
\resizebox{\textwidth}{!}{
\begin{tabular}{clcccl}
\toprule
\textbf{Split} 
& \textbf{Dataset} 
& \textbf{Video Len.} 
& \textbf{Moment Len.} 
& \textbf{Views} 
& \textbf{Domain} \\
\midrule

\multirow{7}{*}{Training}
 & NaQ~\citep{ramakrishnan2023naq}              & 413s & 1.1s  & Ego        & Open \\
 & DiDeMo~\citep{anne2017localizing}            & 29s  & 7.5s  & Exo        & Open \\
 & QuerYD~\citep{oncescu2021queryd}             & 278s & 13.6s & Ego \& Exo & Open \\
 & HiRest~\citep{zala2023hierarchical}          & 263s & 18.9s & Ego \& Exo & Open \\
 & COIN~\citep{tang2019coin}                    & 145s & 14.9s & Ego \& Exo & Open \\
 & Momentor~\citep{qian2024momentor}            & 403s & 49.5s & Ego \& Exo & Open \\
 & YouCook2~\citep{zhou2018towards}             & 316s & 19.7s & Ego \& Exo & Cooking \\
\midrule

\multirow{5}{*}{Benchmark}
 & Ego4D-NLQ~\cite{grauman2022ego4d}            & 500s & 10.7s & Ego        & Open \\
 & TACoS~\cite{regneri2013grounding}            & 368s & 31.9s & Ego \& Exo & Cooking \\
 & Charades-STA~\citep{gao2017tall}             & 29s  & 7.8s  & Ego        & Activity \\
 & ANet-Captions~\cite{krishna2017dense}        & 118s & 40.2s & Ego \& Exo & Activity \\

\bottomrule
\end{tabular}
}
\end{table*}

\section{Dataset Details}
\label{app:dataset_details}
We provide detailed information on the training and evaluation datasets used in our experiments. 
The statistics of each dataset are summarized in Table~\ref{tab:dataset_statistics}.
\paragraph{Training datasets.}
The training set is composed of NaQ~\citep{ramakrishnan2023naq},
DiDeMo~\citep{anne2017localizing}, QuerYD~\citep{oncescu2021queryd},
HiRest~\citep{zala2023hierarchical}, COIN~\citep{tang2019coin},
Momentor~\citep{qian2024momentor}, and YouCook2~\citep{zhou2018towards}.
NaQ provides egocentric temporal grounding annotations derived from episodic-memory queries.
DiDeMo and QuerYD contain natural-language descriptions or narrations aligned with temporal
moments in videos. HiRest provides hierarchical video-moment annotations with step-level
descriptions. COIN and YouCook2 mainly focus on instructional and procedural videos, while
Momentor contains fine-grained temporal reasoning annotations. Together, these datasets provide
a mixture of short and long videos, egocentric and exocentric views, and different query formats.

\paragraph{Training data sampling.}
We do not use all available samples from the above training datasets. Instead, we construct the
hybrid training set by randomly sampling from each dataset with a dataset-specific maximum
sampling budget. Specifically, we sample 8,000 examples from DiDeMo, 5,000 from COIN,
7,000 from YouCook2, 5,000 from QuerYD, 3,500 from HiRest, 7,000 from Momentor, and 7,000 from
NaQ. This sampling strategy balances datasets with different scales and prevents large datasets
from dominating the training process.

\paragraph{Evaluation datasets.}
For short-video temporal grounding, we evaluate on Charades-STA~\citep{gao2017tall}
and ActivityNet-Captions~\citep{krishna2017dense}.
Charades-STA focuses on indoor human activities, ActivityNet-Captions contains diverse
activity videos with dense event captions. For long-video temporal grounding,
we evaluate on TACoS~\citep{regneri2013grounding} and Ego4D-NLQ~\citep{grauman2022ego4d}.
TACoS mainly contains cooking videos with action descriptions, while Ego4D-NLQ consists of
egocentric long videos paired with natural-language queries.

\paragraph{Video question answering benchmarks.}
In addition to temporal grounding benchmarks, we evaluate the general video question answering
ability of our model on MVBench~\citep{li2024mvbench}, 
TempCompass~\citep{liu2403tempcompass}, EgoSchema~\citep{mangalam2023egoschema}, and 
VideoMME~\citep{fu2025video}. These benchmarks cover
different aspects of video understanding, including temporal reasoning, egocentric understanding,
and long-video question answering. For TempCompass, we use all multiple-choice QA tasks except for
the video captioning task. EgoSchema contains egocentric video clips, each approximately
3 minutes long, with temporally demanding QA pairs. VideoMME is a general video QA benchmark
covering diverse domains. It contains 2.7K QA samples over videos of varied lengths, ranging
from 11 seconds to 1 hour. We use the long-video split of VideoMME for evaluation.

Artifact licenses and terms of use.
We use publicly available models and datasets following their released licenses and terms of use. 
Our experiments are conducted for research purposes only, and we do not redistribute the original videos, annotations, or model weights beyond their original access conditions.

\section{Evaluation Metrics}
\label{app:evaluation_metrics}

For temporal grounding evaluation, we report Recall@1 at different IoU thresholds and mean IoU
(mIoU). Given a predicted temporal window $T_{\mathrm{pred}}$ and a ground-truth temporal window
$T_{\mathrm{gt}}$, the temporal Intersection-over-Union is defined as:
\begin{equation}
\mathrm{IoU} =
\frac{|T_{\mathrm{pred}} \cap T_{\mathrm{gt}}|}
{|T_{\mathrm{pred}} \cup T_{\mathrm{gt}}|}.
\end{equation}

Recall@1 at threshold $\theta$, denoted as R@$\theta$, measures whether the top-1 predicted
temporal window has an IoU at least $\theta$ with the ground-truth window:
\begin{equation}
\mathrm{R@}\theta
=
\frac{1}{N}
\sum_{i=1}^{N}
\mathbb{I}
\left[
\operatorname{IoU}
\left(
T_{\mathrm{pred}}^{(i)},
T_{\mathrm{gt}}^{(i)}
\right)
\geq\theta
\right].
\end{equation}

Mean IoU measures the average temporal overlap quality over all queries:
\begin{equation}
\mathrm{mIoU} =
\frac{1}{N}\sum_{i=1}^{N}
\mathrm{IoU}\left(T_{\mathrm{pred}}^{(i)}, T_{\mathrm{gt}}^{(i)}\right),
\end{equation}
where $N$ is the number of evaluated queries.

\subsection{Hyperparameter Selection}

To select the visual chunk size $W^*$ in inference, we evaluated
$W^*\in\{35,40,42,45\}$ using the same checkpoint and a fixed
400-example subset from each of the four temporal-grounding benchmarks.
This resulted in four configurations and 2,000 evaluated examples per
configuration. The corresponding macro-averaged mIoU values were
36.276, 36.924, 37.413, and 36.531, respectively. We therefore selected
$W^*=42$, which is the same as the gradient-gating threshold $W$.

\subsection{Randomness and Reproducibility}

We used a training seed of 42 and a data seed of 42. The Hugging Face training
framework propagates this seed to Python, NumPy, and PyTorch random-number
generators. The duration-aware training sampler uses a dedicated PyTorch
generator initialized with $42+e$ at epoch $e$. The seven-dataset training
mixture was downsampled once using seed 12,345 and stored as a fixed cache,
which was reused by all workers. The cached mixture contains 42,381 examples:
7,000 NAQ, 8,000 DiDeMo, 5,000 QuerYD, 3,381 HiRest,
5,000 COIN, 7,000 Momentor, and 7,000 YouCook2 examples.

\subsection{Computational Infrastructure}

The final model was trained on one node with eight NVIDIA H20 GPUs, each with
97,871 MiB (approximately 96 GB) of device memory. Training used bfloat16
precision, FlashAttention-2, and DeepSpeed ZeRO Stage 3 with both optimizer-state
and parameter offloading to CPU memory. The per-GPU micro-batch size was one and
gradient accumulation was two, giving an effective global batch size of
$8\times1\times2=16$.
The preserved software environment uses Python 3.10.14, PyTorch 2.6.0 with
CUDA 12.4 support, Transformers 4.50.0, TRL 0.15.1, DeepSpeed 0.14.5,
Accelerate 1.4.0, FlashAttention 2.5.8, Datasets 3.3.1, and
Qwen-VL-Utils 0.0.10. The distributed log records NCCL 2.21.5 built for
CUDA 12.4 and CUDA driver API version 12.2. The training framework is our
custom Time-R1~\cite{wang2025time} implementation built on Transformers and TRL.

\subsection{Complete Training Configuration}

We initialized from Qwen2.5-VL-7B-Instruct~\cite{bai2025qwen2}. The vision encoder was frozen and
no parameter-efficient adapter was used. The remaining trainable parameters
were optimized with AdamW using a learning rate of $10^{-6}$, betas
$(0.9,0.999)$, epsilon $10^{-8}$, zero weight decay, and gradient-norm clipping
at 1.0. We used a linear learning-rate schedule without warmup. Training lasted
for two epochs and 5,298 optimizer steps. Checkpoints were saved every 500
steps. Gradient checkpointing and vLLM rollout generation were disabled.

Training rollouts used stochastic decoding with temperature 1.0, top-$p=1.0$,
top-$k=50$, a repetition penalty of 1.0, and a maximum of 32 newly generated
tokens. The configured maximum prompt length was 4,500 tokens. The model used
bfloat16 and FlashAttention-2 throughout training. Videos were decoded at 2 fps.

\begin{table*}[t]
\centering
\small
\setlength{\tabcolsep}{8pt}
\renewcommand{\arraystretch}{1.05}
\begin{tabular}{lcccccc}
\toprule
\textbf{Method}
& \boldmath$T_{\mathrm{vis}}$
& \boldmath$\Delta$\textbf{Mem. (GB)} $\downarrow$
& \textbf{Latency (s)} $\downarrow$
& \textbf{R@0.3} $\uparrow$
& \textbf{R@0.5} $\uparrow$
& \textbf{mIoU} $\uparrow$ \\
\midrule
Uniform-Single
& 3,829
& \textbf{1.50}
& \textbf{1.17}
& 2.49
& 1.33
& 1.77 \\

Full-Dense
& 51,463
& 19.75
& 17.58
& 2.18
& 1.02
& 1.74 \\

SumGround
& 51,463
& 3.95
& 14.78
& \textbf{16.42}
& \textbf{9.51}
& \textbf{11.83} \\
\bottomrule
\end{tabular}
\caption{
Efficiency--accuracy trade-off on Ego4D-NLQ.
$T_{\mathrm{vis}}$ denotes the average cumulative number of visual
tokens processed per sample; for SumGround, it is summed over all
chunk-wise passes.
$\Delta$Mem. denotes the increase in peak GPU memory relative to the
approximately $15.5$~GB model-only footprint measured under the same
inference setup.
}
\label{tab:cost_tradeoff}
\end{table*}

\section{Transfer Evaluation on the TimeLens Benchmark}
\label{app:timelens_benchmark}

To further evaluate the transferability of SumGround beyond its primary long-video setting, we report additional results under the benchmarks proposed by TimeLens~\cite{zhang2025timelens}.
The benchmarks include Charades-TimeLens, ActivityNet-TimeLens, and QVHighlights-TimeLens, which mainly consist of short- to medium-length videos.
Unlike Ego4D-NLQ and TACoS, these datasets do not primarily test the long-horizon evidence identification problem targeted by SumGround.
We therefore use this evaluation to examine whether the proposed long-video summarization framework remains competitive when transferred to shorter temporal contexts.

\begin{table*}[t]
\centering
\small
\setlength{\tabcolsep}{7pt}
\renewcommand{\arraystretch}{1.08}
\begin{tabular}{llcccc}
\toprule
\textbf{Dataset}
& \textbf{Method}
& \textbf{mIoU}
& \textbf{R1@0.3}
& \textbf{R1@0.5}
& \textbf{R1@0.7} \\
\midrule

\multirow{4}{*}{Charades-TimeLens}
& Time-R1
& 36.6
& 57.9
& 32.0
& 16.9 \\
& TimeLens
& \textbf{48.8}
& \textbf{70.5}
& \textbf{55.6}
& \textbf{28.4} \\
& SumGround
& \underline{41.94}
& \underline{62.09}
& \underline{40.11}
& \underline{20.43} \\
\midrule

\multirow{4}{*}{ActivityNet-TimeLens}
& Time-R1
& 33.1
& 44.8
& 31.0
& 19.0 \\
& TimeLens
& \textbf{46.2}
& \textbf{62.8}
& \textbf{51.0}
& \textbf{32.6} \\
& SumGround
& \underline{43.92}
& \underline{59.71}
& \underline{45.88}
& \underline{29.71} \\
\midrule

\multirow{4}{*}{QVHighlights-TimeLens}
& Time-R1
& 49.2
& 65.8
& 51.5
& 36.1 \\
& TimeLens
& 56.0
& \textbf{74.2}
& \textbf{62.7}
& 43.1 \\
& SumGround
& \textbf{56.80}
& \underline{73.78}
& \underline{61.32}
& \textbf{43.15} \\
\bottomrule
\end{tabular}
\caption{
Additional evaluation on the TimeLens benchmark.
Best results are shown in \textbf{bold}, and second-best results are \underline{underlined}.
}
\label{tab:timelens_benchmark}
\end{table*}

As shown in Table~\ref{tab:timelens}, SumGround outperforms Time-R1 across all reported metrics on the three benchmarks.
It ranks second to TimeLens on Charades-TimeLens and ActivityNet-TimeLens.
On QVHighlights-TimeLens, SumGround achieves the best mIoU of 56.80 and slightly exceeds TimeLens on R1@0.7 (43.15 versus 43.10), while remaining competitive on R1@0.3 and R1@0.5.

SumGround is specifically designed to identify query-relevant evidence over extended videos, where dense visual content must be summarized and retrieved across multiple chunks.
The shorter temporal contexts in the TimeLens benchmark place less emphasis on this long-horizon capability and consequently provide limited scope for its main advantage.
Nevertheless, SumGround remains competitive without dataset-specific adaptation, indicating that its long-video-oriented summarization and retrieval design transfers reliably to shorter videos rather than compromising performance outside its primary setting.

\section{Efficiency--Accuracy Trade-off under Long-Video Inference}
\label{app:cost}

We compare different long-video inference strategies in terms of cumulative visual-token processing, peak GPU memory, latency, and temporal-grounding accuracy.
This analysis examines whether the advantage of SumGround comes from processing fewer visual tokens or from organizing dense visual evidence more effectively.

\paragraph{Experimental setting.}
We conduct the analysis on Ego4D-NLQ, which contains long egocentric videos and therefore provides a suitable setting for evaluating long-video inference.
All experiments use bfloat16 precision, greedy decoding, and a maximum of 32 newly generated tokens, and are conducted on a single NVIDIA H20
GPU.
We exclude the first 20 examples as warm-up samples and report the mean end-to-end latency over the remaining examples.

\paragraph{Compared methods.}
We compare three inference strategies:

\begin{itemize}
    \item \textbf{Uniform-Single} uniformly samples a sparse set of frames from the full video such that its visual-token count approximately matches that of one SumGround chunk.
    All sampled frames are processed in a single forward context.

    \item \textbf{Full-Dense} uses approximately the same total visual-token budget as SumGround, but concatenates all sampled frames into a single forward context.

    \item \textbf{SumGround} processes the same dense visual-token budget sequentially across chunks, causally accumulates query-guided summary states, and performs final grounding over associatively retrieved summaries and, when needed, a local visual cache.
\end{itemize}

\paragraph{Metrics.}
$T_{\mathrm{vis}}$ denotes the average cumulative number of visual tokens processed per sample.
For SumGround, this quantity is summed across all chunk-wise passes and therefore measures total visual-token processing rather than the peak number of visual tokens present in a single context.
$\Delta$Mem. denotes the increase in peak GPU memory relative to the approximately $15.5$~GB model-only footprint under the same inference setup.
\textbf{Latency} denotes the mean end-to-end inference time per sample after warm-up.
We report R@0.3, R@0.5, and mIoU for temporal-grounding accuracy.

\paragraph{Discussion.}
As shown in Table~\ref{tab:cost_tradeoff}, Uniform-Single has the lowest inference cost because it processes only 3,829 visual tokens on average.
However, its limited temporal coverage results in poor grounding accuracy.

Full-Dense processes 51,463 visual tokens on average, matching the cumulative visual-token budget of SumGround.
Because all visual tokens are placed in a single context, it incurs substantially higher peak GPU memory.
Moreover, simply increasing the number of visual tokens in a single context does not improve grounding accuracy over Uniform-Single under this setting.

SumGround processes the same cumulative number of visual tokens as Full-Dense while reducing the additional peak GPU memory to an 80.0\% reduction.
It also reduces the average inference latency from 17.58~s to 14.78~s. 
Meanwhile, SumGround improves mIoU from 1.74 to 11.83.

These results indicate that the improvement of SumGround does not arise from processing fewer visual tokens.
Instead, causal summary accumulation and associative retrieval provide a more memory-efficient and effective way to use dense visual evidence for long-video temporal grounding.

\section{Analysis of Associative Retrieval Coverage}
\label{app:selection_coverage}

We evaluate the associative retrieval stage by measuring whether the retrieved chunks cover the ground truth temporal interval before final grounding.
This analysis separates retrieval quality from the final timestamp prediction.

Because each query-guided summary is causally conditioned on preceding summaries, the retrieved summary states may encode evidence originating from earlier chunks.
Therefore, the following metrics measure the explicit temporal coverage of the retrieved source chunks, rather than the complete set of latent evidence available to the grounding decoder.

\paragraph{Evaluation protocol.}
For SumGround, the selected temporal region is defined as the union of
the temporal intervals corresponding to the retrieved summary chunks
and, when used, the source chunk of the auxiliary local visual cache.

For the pretrained Qwen2.5-VL baseline, we use the same chunk
partition, query-association prompts, and associative retrieval rule as
SumGround, but without task-specific RLVR post-training.
This comparison isolates the improvement obtained from learning the
query-association policy.

For UniTime~\cite{li2025universal}, we use its Stage-1 coarse proposal as the selected temporal
region, since this proposal determines the input region used by its
subsequent refinement stage.
All coverage metrics are computed before the final grounding or
refinement step.

\paragraph{Metrics.}
Let $G_n=[t_{s,n},t_{e,n}]$ denote the annotated interval of example
$n$, and let $S_n$ denote its selected temporal region, which may be a
union of multiple chunk intervals.
Over an evaluation set of $N$ examples, we define
\begin{equation}
\mathrm{GT\text{-}Contain}
=
\frac{1}{N}
\sum_{n=1}^{N}
\mathbb{I}
\left[
G_n\subseteq S_n
\right],
\end{equation}
and
\begin{equation}
\mathrm{GT\text{-}Overlap}
=
\frac{1}{N}
\sum_{n=1}^{N}
\mathbb{I}
\left[
G_n\cap S_n\neq\emptyset
\right].
\end{equation}

GT-Contain measures how often the explicitly retrieved source region
covers the complete annotated interval, whereas GT-Overlap measures how
often it intersects at least part of the annotation.
Both metrics are reported as proportions between zero and one.

\begin{table}[t]
\centering
\small
\setlength{\tabcolsep}{4.5pt}
\renewcommand{\arraystretch}{1.05}
\resizebox{\columnwidth}{!}{
\begin{tabular}{lcccc}
\toprule
\multirow{2}{*}{\textbf{Method}}
& \multicolumn{2}{c}{\textbf{Ego4D-NLQ}}
& \multicolumn{2}{c}{\textbf{TACoS}} \\
\cmidrule(lr){2-3}
\cmidrule(lr){4-5}
& \textbf{GT-Contain}
& \textbf{GT-Overlap}
& \textbf{GT-Contain}
& \textbf{GT-Overlap} \\
\midrule
Qwen2.5-VL 
& 0.43 & 0.46
& 0.67 & 0.77 \\

UniTime 
& 0.43 & 0.50
& 0.72 & \textbf{0.89} \\

SumGround
& \textbf{0.52}
& \textbf{0.55}
& \textbf{0.78}
& \underline{0.88} \\
\bottomrule
\end{tabular}}
\caption{
Explicit retrieved-context coverage on long-video temporal grounding
benchmarks.
GT-Contain measures whether the retrieved source region fully contains
the annotated interval, while GT-Overlap measures whether the two have
any temporal intersection.
Values are proportions, and higher is better.
}
\label{tab:selection_coverage}
\end{table}

\paragraph{Discussion.}
As shown in Table~\ref{tab:selection_coverage}, task-specific RLVR
substantially improves the associative retrieval policy over the
pretrained Qwen2.5-VL selector.
On Ego4D-NLQ, SumGround improves GT-Contain from 0.43 to 0.52 and
GT-Overlap from 0.46 to 0.55.
It also outperforms UniTime's Stage-1 proposal on both metrics.
On TACoS, SumGround achieves the highest GT-Contain of 0.78, improving
over UniTime's 0.72, while obtaining a comparable GT-Overlap of 0.88
versus 0.89.
These results indicate that the learned query-association policy more
frequently retrieves chunks covering the complete annotated event,
rather than merely intersecting a small portion of it.

At the same time, the absolute coverage rates reveal an important
remaining bottleneck.
On Ego4D-NLQ, the explicitly retrieved source region fully contains the
annotated interval in only 52\% of the examples; the corresponding rate
on TACoS is 78\%.
The results show that retrieval remains a meaningful system bottleneck.

\section{Effect of the Auxiliary Visual Cache}
\label{app:auxiliary_cache_ablation}

We further examine the contribution of the auxiliary local visual cache
to the final grounding stage.
While the retrieved query-guided summary states provide compact evidence
accumulated across chunks, they may discard fine-grained visual and
temporal cues that are useful for precise boundary prediction.
The auxiliary cache is designed to complement these summaries with
local visual-token states when additional fine-grained evidence is
needed.

\begin{table}[t]
\centering
\small
\setlength{\tabcolsep}{3.5pt}
\renewcommand{\arraystretch}{1.05}
\resizebox{\columnwidth}{!}{
\begin{tabular}{llcccc}
\toprule
\textbf{Dataset}
& \textbf{Variant}
& \textbf{R@0.3}
& \textbf{R@0.5}
& \textbf{R@0.7}
& \textbf{mIoU} \\
\midrule

\multirow{2}{*}{TACoS}
& w/o Auxiliary Cache
& 47.14
& 29.59
& 12.67
& 31.45 \\

& \textbf{SumGround}
& \textbf{52.96}
& \textbf{38.07}
& \textbf{18.67}
& \textbf{35.93} \\

\midrule

\multirow{2}{*}{Charades-STA}
& w/o Auxiliary Cache
& 77.07
& 61.45
& 34.17
& 52.87 \\

& \textbf{SumGround}
& \textbf{79.54}
& \textbf{65.91}
& \textbf{39.57}
& \textbf{56.10} \\

\bottomrule
\end{tabular}
}
\caption{
Ablation of the auxiliary local visual cache on TACoS and
Charades-STA.
The ablated variant sets $\mathcal{C}=\varnothing$ throughout training
and inference, while all other settings remain unchanged.
}
\label{tab:auxiliary_cache_ablation}
\end{table}

\paragraph{Analysis.}
As shown in Table~\ref{tab:auxiliary_cache_ablation}, the auxiliary
cache consistently improves performance across both datasets and all
evaluation metrics.
This pattern suggests that the local visual states are especially useful
for refining temporal boundaries, rather than merely identifying the
coarse location of the queried event.
The results indicate that query-guided summary states and the
auxiliary local visual cache serve complementary roles.
The summary states accumulate query-relevant evidence across chunks,
whereas the auxiliary cache provides fine-grained local cues for precise
boundary prediction.

\section{Qualitative Result}
\begin{figure}[t]
    \centering
    \includegraphics[width=\linewidth]{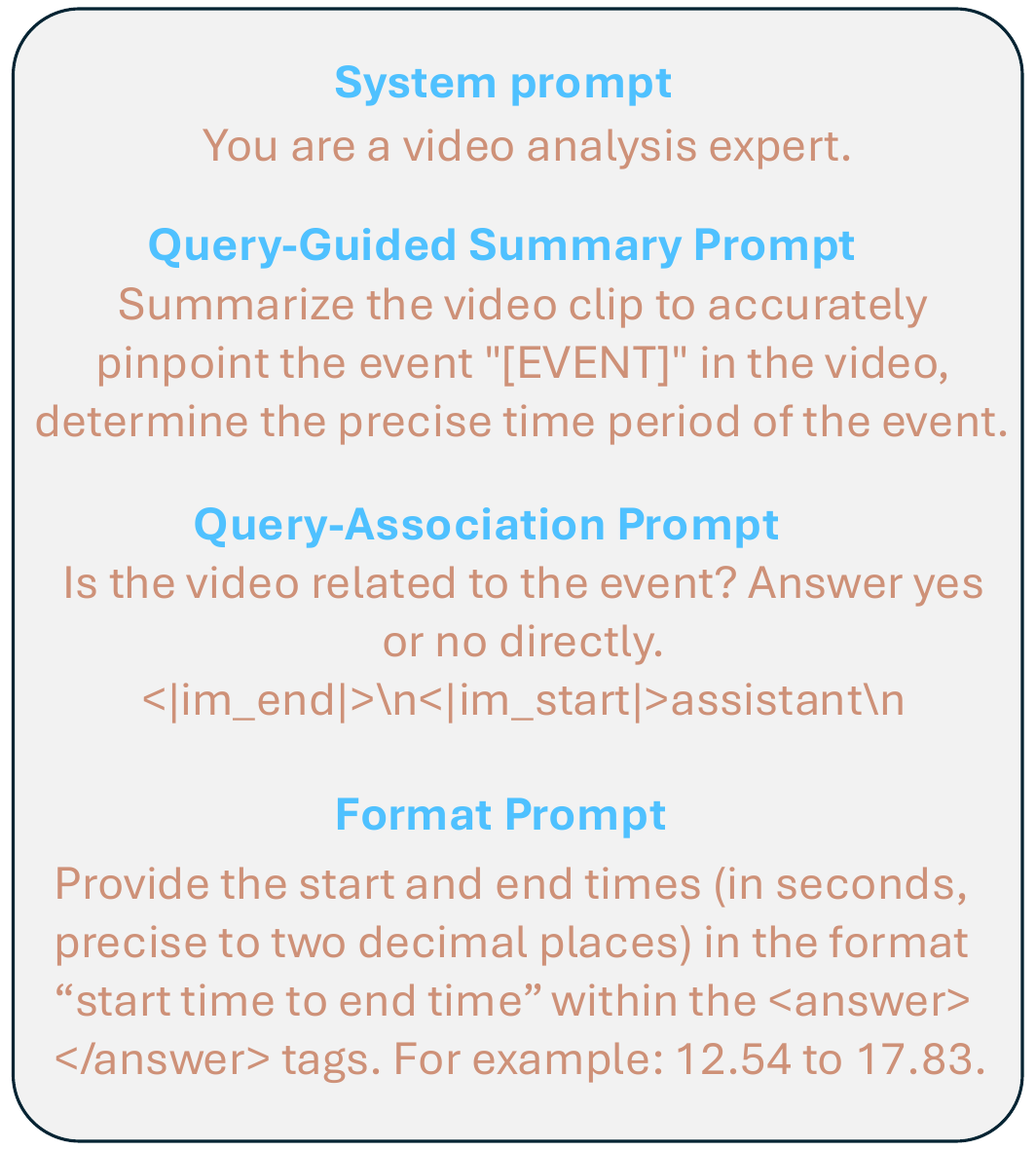}
    \caption{Illustration of prompts at both training and inference time.}
    \label{fig:prompt}
\end{figure}

\paragraph{Illustration of our prompt at training and inference time.} 
Figure~\ref{fig:prompt} presents the prompts used for the temporal video grounding.

\begin{figure}[t]
    \centering
    \small
    \includegraphics[width=\linewidth]{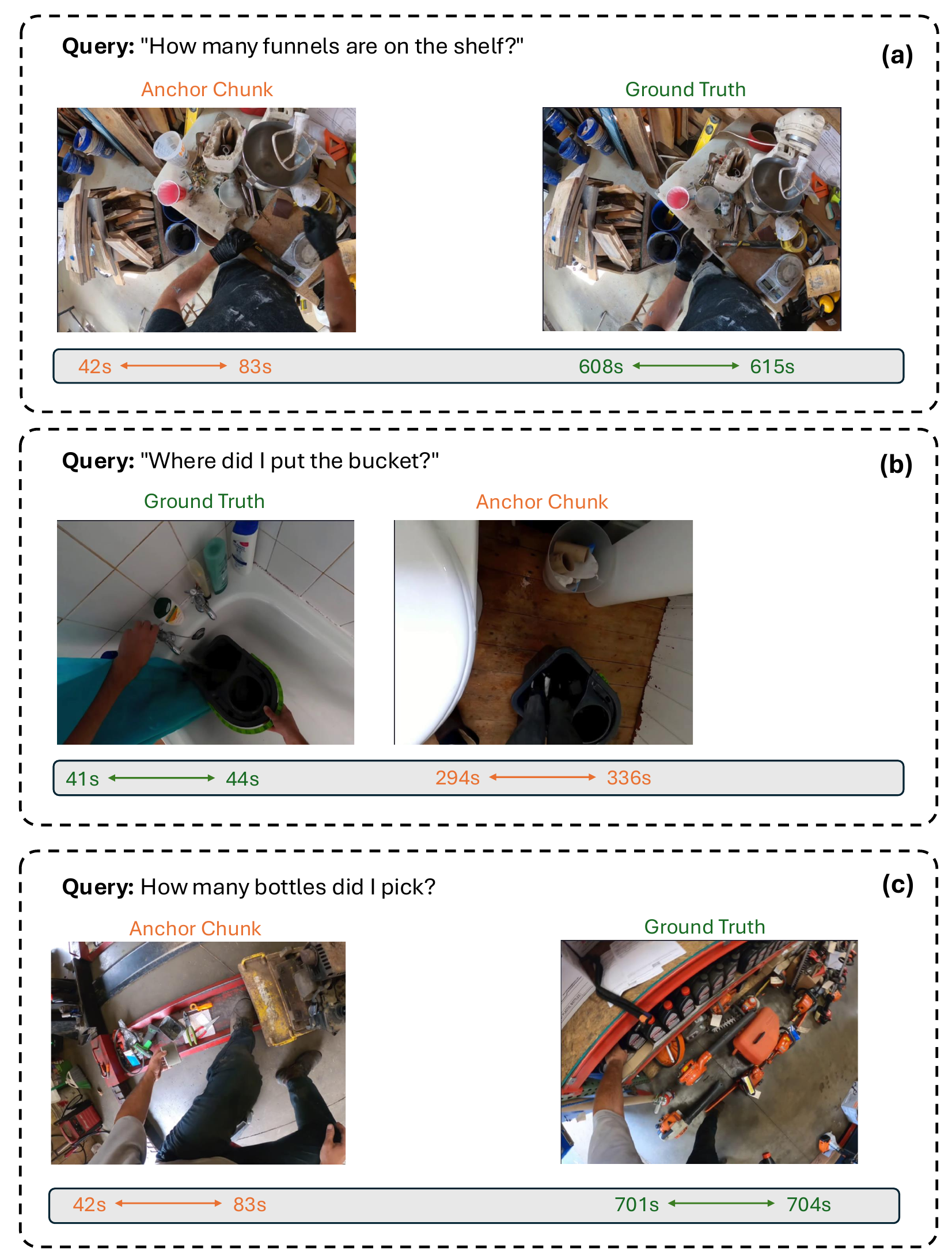}
    \caption{
    Case study of anchor chunk selection on Ego4D-NLQ.
    Orange denotes the selected anchor chunk, while green denotes the annotated ground-truth span.
    Case (c) is a genuine selection error, where the selected chunk contains a similar action but the wrong object.
    In contrast, Cases (a) and (b) show semantically relevant anchor chunks that match the query but fall outside the annotated ground-truth span, suggesting that some measured failures may reflect to repeated or under-annotated query-relevant events.
    }
    \label{fig:case_study}
\end{figure}

\paragraph{Case study on Ego4D-NLQ.}
We conduct a case study on Ego4D-NLQ, the most challenging benchmark in our evaluation, to better understand the failure modes of relevance-guided chunk selection.
Figure~\ref{fig:case_study} shows three examples where the selected anchor chunk does not overlap with the annotated ground-truth span.
Case (c) is a genuine failure: although the selected chunk contains a picking action, the interacted object is not the queried bottle.
In contrast, Cases (a) and (b) reveal a different issue.
The selected anchor chunks are semantically consistent with the queries, as they also show the queried objects or events, but they are not covered by the single annotated ground-truth span.
This suggests that some apparent selection failures are caused by repeated or visually similar query-relevant events outside the annotated interval.
Such cases indicate that evaluation on Ego4D-NLQ may underestimate the quality of relevance-guided selection when multiple valid temporal instances exist in a long egocentric video.

\end{document}